\documentclass[11pt,a4paper]{article}
\usepackage[hyperref]{acl2019}
\usepackage{times}
\usepackage{latexsym}

\usepackage{url}

\usepackage{times}
\usepackage{epsfig}
\usepackage{graphicx}
\usepackage{amsmath}
\usepackage{amssymb}
\usepackage{color}
\usepackage{booktabs}
\usepackage{multirow}

\definecolor{demphcolor}{RGB}{124,124,124}
\newcommand{\demph}[1]{\textcolor{demphcolor}{#1}}
\newcommand{\eg}[0]{e.g.\ }
\newcommand{\ie}[0]{i.e.\ }

\makeatletter
\def\thickhline{%
  \noalign{\ifnum0=`}\fi\hrule \@height \thickarrayrulewidth \futurelet
   \reserved@a\@xthickhline}
\def\@xthickhline{\ifx\reserved@a\thickhline
               \vskip\doublerulesep
               \vskip-\thickarrayrulewidth
             \fi
      \ifnum0=`{\fi}}
\makeatother
\newlength{\thickarrayrulewidth}
\aclfinalcopy 
\def\aclpaperid{292} 

\title{Are You Looking? Grounding to Multiple Modalities in Vision-and-Language Navigation}

\author{Ronghang Hu$^1$, Daniel Fried$^1$, Anna Rohrbach$^1$, Dan Klein$^1$, Trevor Darrell$^1$, Kate Saenko$^2$ \\
$^1$University of California, Berkeley $\qquad$ $^2$Boston University \\
\texttt{\{ronghang,dfried,anna.rohrbach,klein\}@cs.berkeley.edu,}\\
\texttt{trevor@eecs.berkeley.edu,saenko@bu.edu} \\}

\date{}

\begin{document}
\maketitle

\begin{abstract}
Vision-and-Language Navigation (VLN) requires grounding instructions, such as \emph{turn right and stop at the door}, to routes in a visual environment. The actual grounding can connect language to the environment through multiple modalities, \eg \emph{stop at the door} might ground into visual objects, while \emph{turn right} might rely only on the geometric structure of a route. We investigate \emph{where} the natural language empirically grounds under two recent state-of-the-art VLN models. Surprisingly, we discover that visual features may actually \emph{hurt} these models: models which only use route structure, ablating visual features, outperform their visual counterparts in unseen new environments on the benchmark Room-to-Room dataset. To better use all the available modalities, we propose to decompose the grounding procedure into a set of expert models with access to different modalities (including object detections) and ensemble them at prediction time, improving the performance of state-of-the-art models on the VLN task.
\end{abstract}

\section{Introduction}\label{sec:intro} 

The Vision-and-Language Navigation (VLN) task \cite{anderson2018cvpr} requires an agent to navigate to a particular location in a real-world environment, following complex, context-dependent instructions written by humans (\eg \textit{go down the second hallway on the left, enter the bedroom and stop by the mirror}). The agent must navigate through the environment, conditioning on the instruction as well as the visual imagery that it observes along the route, to stop at the location specified by the instruction (\eg \emph{the mirror}). 

\begin{figure}[t]
\vspace{-1em}
\small
\centering
\includegraphics[width=0.95\linewidth]{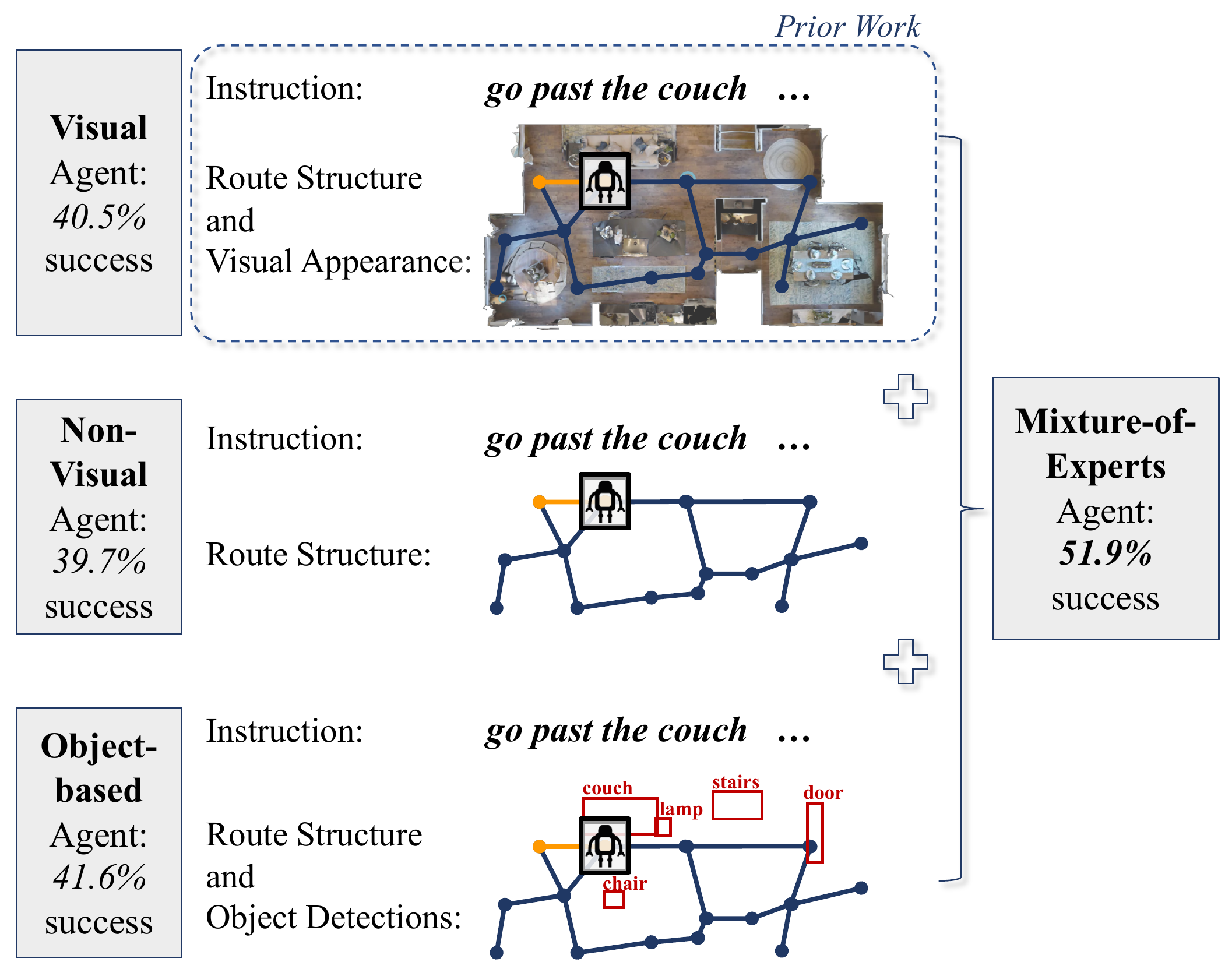}
\vspace{-.5em}
\caption{We factor the grounding of language instructions into visual appearance, route structure, and object detections using a mixture-of-experts approach.}
\label{fig:approach}
\vspace{-2em}
\end{figure}

Recent state-of-the-art models 
\cite{wang2018look,fried2018speaker,anonymous2019self-aware} have demonstrated large gains in accuracy on the VLN task. However, it is unclear which \textit{modality} these substantial increases in task metrics can be attributed to, and, in particular, whether the gains in performance are due to stronger grounding into visual context or \eg simply into the discrete, geometric structure of possible routes, such as turning left or moving forward (see Fig.~\ref{fig:approach}, top vs. middle).

First, we analyze to what extent VLN models ground language into visual appearance and route structure by training versions of two state-of-the-art models \emph{without} visual features, using the benchmark Room-to-Room (R2R) dataset \cite{anderson2018cvpr}. We find that while grounding into route structure is useful, the models with visual features fail to learn generalizable visual grounding. Surprisingly, when trained without visual features, their performance on unseen environments is comparable or even better.

We hypothesize that the low-level, pixel-based CNN features in the visual models contribute to their failure to generalize. To address this, 
we introduce a high-level object-based visual representation to ground language into visual context in a more generalizable way, using the symbolic output of a pretrained object detection system. For example, while a concept \emph{table} could ground into visual appearance of a specific table in a given environment, detecting tables and other objects in scenes, mapping them into symbols, and grounding the text mentions into these symbols should generalize better to unseen environments.

Finally, inspired by the complementary errors of visual and non-visual agents, we decompose the grounding process through a mixture-of-experts approach.
We train separate visual and non-visual agents, encouraging each one to focus on a separate modality, and combine their predictions as an ensemble (see Fig.~\ref{fig:approach}). Our mixture-of-experts outperforms the individual agents, and is also better than the ensembles of multiple agents of the same modality (\eg both visual or both non-visual).

Adding our object representation and mixture-of-experts approach to both state-of-the-art models improves their success rate by over 10\% (absolute) in novel environments, 
obtaining a 51.9\% success rate on the \emph{val-unseen} split of the benchmark R2R dataset \cite{anderson2018cvpr}.

\section{Related work}\label{sec:related}

\paragraph{Vision and Language Navigation.}
Vision-and-Language Navigation (VLN) \cite{anderson2018cvpr,chen2019touchdown} 
unites two lines of work: first, of following natural language navigational instructions in an environmental context \cite{macmahon2006walk,vogel2010learning,tellex2011understanding,chen2011navigation,artzi2013instructions,Andreas15Instructions,Mei16Instructions,fried2017unified,misra2018mapping}, and second, of vision-based navigation tasks  \cite{mirowski2016learning,zhu2017target,yang2018visual,mirowski2018learning,cirik2018following}
that use visually-rich real-world imagery \cite{chang2017matterport3d}.

A number of methods for the VLN task have been recently proposed. \newcite{wang2018look} use model-based and model-free reinforcement learning to learn an environmental model and optimize directly for navigation success. \newcite{fried2018speaker} use a separate instruction generation model to synthesize new instructions as data augmentation during training, and perform pragmatic inference at test time. Most recently, \newcite{anonymous2019self-aware} introduce a visual and textual co-attention mechanism and a route progress predictor.

These approaches have significantly improved performance on the VLN task, when evaluated by metrics such as \emph{success rate}. However, it is unclear where the high performance comes from. In this paper, we find that agents without any visual input can achieve competitive performance, matching or even outperforming their vision-based counterparts under two state-of-the-art model models \cite{fried2018speaker,anonymous2019self-aware}. We also explore two approaches to make the agents better utilize their visual inputs.

\paragraph{The role of vision in vision-and-language tasks.}
In several vision-and-language tasks, high performance can be achieved without effective modeling of the visual modality. 
\newcite{DBLP_conf_acl_DevlinCFGDHZM15} find that image captioning models can exploit regularity in the captions, showing that a nearest-neighbor matching approach can achieve competitive performance to sophisticated language generation models. \newcite{hendricks2018women} and \newcite{rohrbach2018emnlp} find that neural captioning models often ground object mentions into incorrect objects due to correlations in the training data, and can hallucinate non-existing objects.

Recent work has also investigated single-modality performance in vision-and-language embodiment tasks. \newcite{anand2018blindfold} find that state-of-the-art results can be achieved on the EmbodiedQA task \cite{das2017embodied} using an agent without visual inputs. Work concurrent to ours evaluates the performance of single-modality models for several embodied tasks including VLN \cite{thomason2018shifting}, finding that high performance can be achieved on the R2R dataset using a non-visual version of the baseline model \cite{anderson2018cvpr}. 
In this paper, we show that the same trends hold for two recent state-of-the-art architectures \cite{anonymous2019self-aware,fried2018speaker} 
for the VLN task; we also analyze to what extent object-based representations and mixture-of-experts methods can address these issues.

\section{State-of-the-art VLN models do not use vision effectively}\label{sec:nonvision_analysis}

We experiment with the benchmark Room-to-Room (R2R) dataset \cite{anderson2018cvpr} for the Vision-and-Language navigation task, which consists of a set of annotated instructions for routes through \emph{environments} from the Matterport3D dataset \cite{chang2017matterport3d}. Each environment is a building, such as a house or office, containing a set of \emph{viewpoints}: physical locations in the environment, each with an associated panoramic image. Viewpoints are connected in a connectivity graph determined by line-of-sight in the physical environment. 
See the top row of Fig.~\ref{fig:approach} for a top-down environment illustration.

In the VLN task, a virtual agent is placed at a particular viewpoint in an environment, and is given a natural language instruction (written by a human annotator) to follow. At each timestep, the agent receives the panoramic image for the viewpoint it is currently located at, and either predicts to move to one of the adjacent connected viewpoints, or to stop. When the agent predicts the stop action, it is evaluated on whether it has correctly reached the end of the route that the human annotator was asked to describe. 

In this work, we analyze two recent VLN models, which typify the visual grounding approaches of VLN work: the panoramic ``follower'' model from the Speaker-Follower (SF) system of \citet{fried2018speaker} and the Self-Monitoring (SM) model of \citet{anonymous2019self-aware}.
These models obtained state-of-the-art results on the R2R dataset.
Both models are based on the encoder-decoder approach \cite{cho2014properties} and map an instruction to a sequence of actions in context by encoding the instruction with an LSTM, and outputting actions using an LSTM decoder that conditions on the encoded instruction and visual features summarizing the agent's environmental context. 
Compared to the SF model, the SM model introduces an improved visual-textual co-attention mechanism and a progress monitor component.
We refer to the original papers for details on the two models.

To analyze the models' visual grounding ability, we focus on their core encoder-decoder components. In our experiments, we use models trained without data augmentation, and during inference predict actions with greedy search (\ie without beam search, pragmatic, or progress monitor-based inference). For SF, we use the publicly released code. For SM, we use a reimplementation without the progress monitor, which was shown to be most important for search in inference \cite{anonymous2019self-aware}.

\begin{table}[t]
\scriptsize
\begin{center}
\begin{tabular}{c|c|c|c|c|c}
\toprule
& model & train. & vis. & SR on & SR on \\
\# & arch. & appr. & feat. & val-seen & val-unseen \\
\hline
1 & \multirow{4}{*}{SF} & \multirow{2}{*}{stud.-forc.}
& no vis. & \demph{29.7} & \textbf{31.7} \\
2 & \multirow{3}{*}{} &
& RN & \demph{53.3} & 29.0 \\
\cline{3-6}
3 & & \multirow{2}{*}{teach.-forc.}
& no vis. & \demph{34.1} & \textbf{35.2} \\
4 & & 
& RN & \demph{40.4} & 29.0  \\
\hline
5 & \multirow{4}{*}{SM} & \multirow{2}{*}{stud.-forc.}
& no vis. & \demph{36.1} & 39.7 \\
6 & \multirow{3}{*}{} &
& RN & \demph{62.8} & \textbf{40.5}\\
\cline{3-6}
7 & & \multirow{2}{*}{teach.-forc.}
& no vis. & \demph{34.3} & 32.2 \\
8 & &
& RN & \demph{44.0} & \textbf{32.8} \\
\bottomrule
\end{tabular}
\end{center}
\vspace{-1em}
\caption{Success rate (SR) of the vision-based full agent (``RN'', using ResNet) and the non-visual agent (``no vis.'', setting all visual features to zero) on the R2R dataset under different model architectures (Speaker-Follower (SF)~\cite{fried2018speaker} and Self-Monitoring (SM)~\cite{anonymous2019self-aware}) and training schemes.}
\label{tab:vis_nonvis}
\vspace{-2em}
\end{table}

We investigate how well these models ground instructions into visual features of the environment, by training and evaluating them 
without access to the visual context:
setting their visual feature vectors to 
zeroes during training and testing. 
We compare performance on the validation sets of the R2R dataset: 
the \emph{val-seen} split, consisting of the same environments as in training, and the \emph{val-unseen} split of novel environments. Since we aim to evaluate how well the agents \textit{generalize} to the unseen environments, we focus on the val-unseen split.
For both the SF and SM models, we train two versions of the agents, using either the student-forcing or teacher-forcing approaches of \citet{anderson2018cvpr}\footnote{During training, the agent either follows the ground-truth actions (teacher-forcing) or samples actions from its own prediction (student-forcing). See supplemental for more details.}, and select the best training snapshot on the val-seen split.\footnote{Following previous work, we use success rate (SR) as our evaluation metric, where an episode is considered successful if the agent stops within 3 meters of the goal location.} The results are shown in Table~\ref{tab:vis_nonvis}. In each block, the two rows show the agent's performance (under the specific model architecture and training approach) with or without access to the visual features (``RN'': ResNet-152 network \cite{he2016deep}, ``no vis.'': non-visual). 

While visual features improve performance on environments seen during training, 
we see that for the SF architecture the non-visual agent (lines 1 and 3) outperforms the visual agent (lines 2 and 4) on unseen environments under both student-forcing and teacher-forcing training. For SM, the non-visual agent (lines 5 and 7) has a success rate very close to the visual agent (lines 6 and 8). 
This indicates that these models do not learn generalizable visual perception, so that the visual features may actually \textit{hurt} them in unseen environments. 

\section{Object representation for better grounding and generalization}\label{sec:exp_object}

In both the SF and SM architectures, the agents use visual features from a pretrained ResNet-152 CNN \cite{he2016deep}. 
As the training data for the R2R dataset contains only 61 distinct environments, the agents may overfit to the appearance of the training environments and thus struggle to generalize. 
For example, for the instruction \emph{go down the staircase}, a model may learn to ground \emph{staircase} into a specific staircase in a given training environment, and fail to generalize to staircases with different appearances or in different contexts in unseen environments. 
We thus propose an \textit{object-based representation}, where object detection results from a pretrained large-scale object detector are used as the environment representation. The object-based representation is intended to prevent overfitting to training scenes and to transfer to new environments better than CNN features.

Both the SF and SM models represent the visual appearance at each location 
with a set of visual features $\{x_{img,i}\}$, where $x_{img,i}$ is a vector extracted from an image patch at a particular orientation $i$ using a CNN. Both models also use a visual attention mechanism to extract an attended visual feature $x_{img,att}$ from $\{x_{img,i}\}$. 
For our object-based representation, we use a Faster R-CNN \cite{ren2015faster} object detector trained on the Visual Genome dataset \cite{krishna2017visual}. 
We construct a set of vectors $\{x_{obj,j}\}$ representing detected objects and their attributes. Each vector $x_{obj,j}$ ($j$-th detected object in the scene) is a concatenation of summed GloVe vectors \cite{pennington2014glove} for the detected object label (\textit{e.g.\ }\emph{door}) and attribute labels (\textit{e.g.\ }\emph{white}) and a location vector from the object's bounding box coordinates. 
We then use the same visual attention mechanism as in \newcite{fried2018speaker} and \newcite{anonymous2019self-aware} to obtain an attended object representation $x_{obj,att}$ over these $\{x_{obj,j}\}$ vectors. 
We either substitute the ResNet CNN features $x_{img,att}$ (``RN'') with our object representation $x_{obj,att}$ (``Obj''), or concatenate $x_{img,att}$ and $x_{obj,att}$ (``RN+Obj''). Then we train the SF model or the SM model using this object representation, with results shown in Table~\ref{tab:objects}.\footnote{For each model and setting, we use the best training mechanism as found in Table~\ref{tab:vis_nonvis}, where student-forcing is used in all experiments except line 1 (where teacher-forcing is used to obtain the best performance for the non-visual agent under the SF architecture). See supplemental for more details.} 

For SF (lines 1--4), object representations substantially improve generalization ability: using either the object representation (``Obj'') or the combined representation (``RN+Obj'') obtains higher success rate on unseen environments than using only the ResNet features (``RN''), and the combined representation (``RN+Obj'') obtains the highest overall performance. 
For SM (lines 5--8), the model that uses only the object representation achieves the best performance (line 7). Here the success rates across the four settings are closer, and the improvement from object representation is smaller than for SF. However, in Sec.~\ref{sec:exp_moe} we find that object representation can be combined with other inputs to further improve the performance. 

\begin{table}[t]
\scriptsize
\begin{center}
\begin{tabular}{c|c|c|c|c}
\toprule
& model & vis. & SR on & SR on \\
\# & arch. & feat. & val-seen & val-unseen\\
\hline
1 & \multirow{4}{*}{SF}
& no vis. & \demph{34.1}  & 35.2 \\
2 & \multirow{3}{*}{}
& RN & \demph{53.3} & 29.0 \\
3 & 
& Obj & \demph{38.5} & 33.5 \\
4 & 
& RN+Obj & \demph{47.8} & \textbf{39.8} \\
\hline
5 & \multirow{4}{*}{SM}
& no vis. & \demph{36.1} & 39.7 \\
6 & \multirow{3}{*}{}
& RN & \demph{62.8} & 40.5  \\
7 & 
& Obj & \demph{48.8} & \textbf{41.6}\\
8 & 
& RN+Obj & \demph{59.2} & 39.5 \\
\bottomrule
\end{tabular}
\end{center}
\vspace{-1.5em}
\caption{Success rate (SR) of agents with different visual inputs on the R2R dataset (``RN'': ResNet CNN, ``Obj'': objects, ``no vis.'': no visual representation). Models: Speaker-Follower (SF)~\cite{fried2018speaker} and Self-Monitoring (SM)~\cite{anonymous2019self-aware}.}
\label{tab:objects}
\vspace{-1.5em}
\end{table}

\section{Mixture-of-experts makes better use of all available information}\label{sec:exp_moe}

While the agent with CNN visual features does not outperform its non-visual counterpart (Sec.~\ref{sec:nonvision_analysis}) on average, it often succeeds on individual instructions where the non-visual model fails, indicating the visual and non-visual modalities are complementary. 
To encourage grounding into both modalities, we ensemble visual and non-visual models in a mixture-of-experts approach.

\subsection{Separate Training}
\label{sec:separate_training}
We first ensemble the models from Sec.~\ref{sec:nonvision_analysis} and Sec.~\ref{sec:exp_object} at test time (after training them separately) by combining their predictions at each timestep.\footnote{We combine model predictions at each timestep by averaging action logits across models, which in early experiments slightly outperformed averaging action probabilities.}

\begin{table}
\scriptsize
\begin{center}
\begin{tabular}{c|c|c|c}
\toprule
& model & mix.-of-exp. & SR on \\
\# & arch. & comb. & val-unseen \\

\hline
9 & \multirow{5}{*}{SF}
& (no vis., no vis.) & 35.1
\\
10 &
& (RN, RN) & 32.1
\\
11 & \multirow{3}{*}{(mixture}
& (Obj, Obj) & 35.4
\\
12 & \multirow{3}{*}{of 2 models)}
& (RN+Obj, RN+Obj) & \textbf{43.3}
\\
13 & 
& (RN, no vis.) & 39.5
\\
14 & 
& (Obj, no vis.) & 38.4
\\
15 & 
& (RN+Obj, no vis.) & 43.1
\\
\thickhline
16 & \multirow{5}{*}{SM}
& (no vis., no vis.) & 41.0
\\
17 &
& (RN, RN) & 43.5 
\\
18 & \multirow{3}{*}{(mixture}
& (Obj, Obj) & 45.2
\\
19 & \multirow{3}{*}{of 2 models)}
& (RN+Obj, RN+Obj) & 42.2 
\\
20 & 
& (RN, no vis.) & \textbf{46.9}
\\
21 & 
& (Obj, no vis.) & 43.4 
\\
22 & 
& (RN+Obj, no vis.) & 46.4
\\
\hline
23 & SM (3-way mix.)
& (RN, Obj, no vis.) & \textbf{49.5} 
\\
\hline
24 & SM
& (RN, no vis.) & 48.3
\\
25 & \multirow{1}{*}{(joint training)}
& (RN, Obj, no vis.) & \textbf{51.9} 
\\
\bottomrule
\end{tabular}
\end{center}
\vspace{-1em}
\caption{Success rate (SR) of different mixture-of-experts ensembles. 
Models: Speaker-Follower (SF)~\cite{fried2018speaker} and Self-Monitoring (SM)~\cite{anonymous2019self-aware}; ``RN'': ResNet CNN, ``Obj'': objects, ``no vis.'': no visual representation.}
\label{tab:moe}
\vspace{-2em}
\end{table}

Lines 9--22 of Table~\ref{tab:moe} show ensembles of two 
models. 
Compared to single-model performance (line 1--8 in Table~\ref{tab:objects}), an ensemble of a visual and a non-visual agent outperforms the individual agents for both the SF and the SM models. The best performing setting is the combination of ``RN'' and ``no vis.'' (non-visual) in line 20 under the SM model. 
While it is unsurprising that the mixture-of-experts can boost performance, it is interesting to see that the best 
mixture in line 20 outperforms mixtures of two agents of the same type (two non-visual agents in line 16, two visual agents in line 17, trained from distinct random parameter initializations), confirming 
that two agents with access to different modalities can complement 
each other, especially in the SM model.

We also experiment with a 3-way mixture in the SM model, combining a visual agent with ResNet CNN features, a visual agent with object features, 
and a non-visual agent (line 23). This mixture outperforms all the 2-way mixtures by a noticeable margin, showing that the CNN and object-based visual representations are also complementary.

\subsection{Joint Training}
\label{sec:joint_training}
Finally, given the success of this simple test-time ensemble, we also explore \textit{jointly training} these models by building a \textit{single agent} which uses a single instruction encoder shared between multiple (visual and non-visual) jointly-trained decoders. 
During joint training, each decoder is supervised to predict the true actions, applying the same loss function as in separate training.
During testing, actions are predicted by averaging logits from the separate decoders as in Sec.~\ref{sec:separate_training}. 
We experiment with jointly training the agents in each of the two best-performing combinations (RN, no vis.) and (RN, Obj, no vis.) under the SM architecture (line 24 and 25 of Table~\ref{tab:moe}). From line 24 vs.\ 20 and line 25 vs.\ 23, joint training gives higher performance than training each model separately and combining them only at test time. 
Overall, we obtain 51.9\% final success rate on the val-unseen split (line 25), 
which is over 10\% (absolute) higher than the SF or SM baselines using a single decoder with CNN features (line 2 and 6 in Table~\ref{tab:objects}).\footnote{\label{ftn:val_seen}On the val-seen split, we maintain performance comparable to the original SM model.}

\section{Discussion}

The success of non-visual versions of two recent state-of-the-art VLN models, 
often outperforming their vision-based counterparts in unseen environments on the benchmark R2R dataset, shows that these models 
do not use the visual inputs in a generalizable way. 
Our intuition is that, while language has rich, high-level symbolic meaning, which can be easily matched to the modality of the route structures, pixel-based visual representations, even those extracted via CNNs, are a lower-level modality which require more data to learn, and so a model trained on both modalities may learn to mostly rely on the route structure. 
This is also supported by the results in Table~\ref{tab:moe} (line 23 vs. line 20), where adding higher-level object representations improves the success rate by 2.6\%. 

Notably, an agent in the R2R environment is only able to move to a discrete set of locations in the environment, and at each point in time it only has a small number of actions available, determined by the environment's connectivity graph (i.e., moving to the adjacent locations). 
These constraints on possible routes help explain our findings that language in the VLN instructions often grounds into geometric route structure in addition to visual context along the route. For example, if an instruction says \textit{turn left at the couch}, and the route structure only allows the agent to turn left at a single location, it may not need to perceive the couch. Other instructions, such as \textit{go straight for 5 meters and stop} may also be carried out  without access to visual perception.

The improvement of our mixture-of-experts approach 
over single models suggests that it is challenging to learn to ground language into multiple modalities in one model. The ``RN+Obj'' model (Table~\ref{tab:objects}, line~8) has access to \emph{the same} information as our best result in Table~\ref{tab:moe}, line~25, but obtains much lower success rate (39.5\% vs. 51.9\%). 
Thus, splitting the prediction task across several models, where each has access to a different input modality, is an effective way to inject an inductive bias that encourages the model to ground into each of the modalities.

\paragraph{Acknowledgements.} This work was partially supported by Berkeley AI
Research, the NSF and DARPA XAI. 
DF is supported by a Tencent AI Lab Fellowship.

\bibliographystyle{acl_natbib}
\bibliography{biblioLong,acl2019}

\begin{thebibliography}{33}
\expandafter\ifx\csname natexlab\endcsname\relax\def\natexlab#1{#1}\fi

\bibitem[{Anand et~al.(2018)Anand, Belilovsky, Kastnerand, Larochelle, and
  Courville}]{anand2018blindfold}
Ankesh Anand, Eugene Belilovsky, Kyle Kastnerand, Hugo Larochelle, and Aaron
  Courville. 2018.
\newblock Blindfold baselines for embodied qa.
\newblock \emph{arXiv preprint arXiv:1811.05013}.

\bibitem[{Anderson et~al.(2018)Anderson, Wu, Teney, Bruce, Johnson,
  S{\"u}nderhauf, Reid, Gould, and Hengel}]{anderson2018cvpr}
Peter Anderson, Qi~Wu, Damien Teney, Jake Bruce, Mark Johnson, Niko
  S{\"u}nderhauf, Ian Reid, Stephen Gould, and Anton van~den Hengel. 2018.
\newblock Vision-and-language navigation: Interpreting visually-grounded
  navigation instructions in real environments.
\newblock In \emph{Proceedings of the IEEE Conference on Computer Vision and
  Pattern Recognition (CVPR)}.

\bibitem[{Andreas and Klein(2015)}]{Andreas15Instructions}
Jacob Andreas and Dan Klein. 2015.
\newblock Alignment-based compositional semantics for instruction following.
\newblock In \emph{Proceedings of the Conference on Empirical Methods in
  Natural Language Processing (EMNLP)}.

\bibitem[{Artzi and Zettlemoyer(2013)}]{artzi2013instructions}
Yoav Artzi and Luke Zettlemoyer. 2013.
\newblock Weakly supervised learning of semantic parsers for mapping
  instructions to actions.
\newblock \emph{Transactions of the Association for Computational Linguistics},
  1(1):49--62.

\bibitem[{Bahdanau et~al.(2015)Bahdanau, Cho, and Bengio}]{bahdanau2014neural}
Dzmitry Bahdanau, Kyunghyun Cho, and Yoshua Bengio. 2015.
\newblock Neural machine translation by jointly learning to align and
  translate.
\newblock In \emph{Proceedings of the International Conference on Learning
  Representations (ICLR)}.

\bibitem[{Chang et~al.(2017)Chang, Dai, Funkhouser, Halber, Niebner, Savva,
  Song, Zeng, and Zhang}]{chang2017matterport3d}
Angel Chang, Angela Dai, Thomas Funkhouser, Maciej Halber, Matthias Niebner,
  Manolis Savva, Shuran Song, Andy Zeng, and Yinda Zhang. 2017.
\newblock Matterport3d: Learning from rgb-d data in indoor environments.
\newblock In \emph{2017 International Conference on 3D Vision (3DV)}, pages
  667--676. IEEE.

\bibitem[{Chen and Mooney(2011)}]{chen2011navigation}
David~L. Chen and Raymond~J. Mooney. 2011.
\newblock Learning to interpret natural language navigation instructions from
  observations.
\newblock In \emph{Proceedings of the Conference on Artificial Intelligence
  (AAAI)}.

\bibitem[{Chen et~al.(2019)Chen, Shur, Misra, Snavely, and
  Artzi}]{chen2019touchdown}
Howard Chen, Alane Shur, Dipendra Misra, Noah Snavely, and Yoav Artzi. 2019.
\newblock Touchdown: Natural language navigation and spatial reasoning in
  visual street environments.
\newblock In \emph{Proceedings of the IEEE Conference on Computer Vision and
  Pattern Recognition (CVPR)}.

\bibitem[{Cho et~al.(2014)Cho, Van~Merri{\"e}nboer, Bahdanau, and
  Bengio}]{cho2014properties}
Kyunghyun Cho, Bart Van~Merri{\"e}nboer, Dzmitry Bahdanau, and Yoshua Bengio.
  2014.
\newblock On the properties of neural machine translation: Encoder-decoder
  approaches.
\newblock \emph{arXiv preprint arXiv:1409.1259}.

\bibitem[{Cirik et~al.(2018)Cirik, Zhang, and Baldridge}]{cirik2018following}
Volkan Cirik, Yuan Zhang, and Jason Baldridge. 2018.
\newblock Following formulaic map instructions in a street simulation
  environment.
\newblock In \emph{Visually Grounded Interaction and Language (ViGIL) Workshop,
  NeurIPS}.

\bibitem[{Das et~al.(2018)Das, Datta, Gkioxari, Lee, Parikh, and
  Batra}]{das2017embodied}
Abhishek Das, Samyak Datta, Georgia Gkioxari, Stefan Lee, Devi Parikh, and
  Dhruv Batra. 2018.
\newblock Embodied question answering.
\newblock In \emph{Proceedings of the IEEE Conference on Computer Vision and
  Pattern Recognition (CVPR)}.

\bibitem[{Devlin et~al.(2015)Devlin, Cheng, Fang, Gupta, Deng, He, Zweig, and
  Mitchell}]{DBLP_conf_acl_DevlinCFGDHZM15}
Jacob Devlin, Hao Cheng, Hao Fang, Saurabh Gupta, Li~Deng, Xiaodong He,
  Geoffrey Zweig, and Margaret Mitchell. 2015.
\newblock Language models for image captioning: The quirks and what works.
\newblock In \emph{Proceedings of the 53rd Annual Meeting of the Association
  for Computational Linguistics and the 7th International Joint Conference on
  Natural Language Processing of the Asian Federation of Natural Language
  Processing, {ACL} 2015, July 26-31, 2015, Beijing, China, Volume 2: Short
  Papers}.

\bibitem[{Fried et~al.(2018{\natexlab{a}})Fried, Andreas, and
  Klein}]{fried2017unified}
Daniel Fried, Jacob Andreas, and Dan Klein. 2018{\natexlab{a}}.
\newblock Unified pragmatic models for generating and following instructions.
\newblock In \emph{Proceedings of the Conference of the North American Chapter
  of the Association for Computational Linguistics (NAACL)}.

\bibitem[{Fried et~al.(2018{\natexlab{b}})Fried, Hu, Cirik, Rohrbach, Andreas,
  Morency, Berg-Kirkpatrick, Saenko, Klein, and Darrell}]{fried2018speaker}
Daniel Fried, Ronghang Hu, Volkan Cirik, Anna Rohrbach, Jacob Andreas,
  Louis-Philippe Morency, Taylor Berg-Kirkpatrick, Kate Saenko, Dan Klein, and
  Trevor Darrell. 2018{\natexlab{b}}.
\newblock Speaker-follower models for vision-and-language navigation.
\newblock In \emph{Advances in Neural Information Processing Systems (NIPS)}.

\bibitem[{He et~al.(2016)He, Zhang, Ren, and Sun}]{he2016deep}
Kaiming He, Xiangyu Zhang, Shaoqing Ren, and Jian Sun. 2016.
\newblock Deep residual learning for image recognition.
\newblock In \emph{Proceedings of the IEEE Conference on Computer Vision and
  Pattern Recognition (CVPR)}, pages 770--778.

\bibitem[{Hendricks et~al.(2018)Hendricks, Burns, Saenko, Darrell, and
  Rohrbach}]{hendricks2018women}
Lisa~Anne Hendricks, Kaylee Burns, Kate Saenko, Trevor Darrell, and Anna
  Rohrbach. 2018.
\newblock Women also snowboard: Overcoming bias in captioning models.
\newblock In \emph{European Conference on Computer Vision}, pages 793--811.
  Springer.

\bibitem[{Hochreiter and Schmidhuber(1997)}]{hochreiter1997long}
Sepp Hochreiter and J{\"u}rgen Schmidhuber. 1997.
\newblock Long short-term memory.
\newblock \emph{Neural computation}, 9(8):1735--1780.

\bibitem[{Krishna et~al.(2017)Krishna, Zhu, Groth, Johnson, Hata, Kravitz,
  Chen, Kalantidis, Li, Shamma et~al.}]{krishna2017visual}
Ranjay Krishna, Yuke Zhu, Oliver Groth, Justin Johnson, Kenji Hata, Joshua
  Kravitz, Stephanie Chen, Yannis Kalantidis, Li-Jia Li, David~A Shamma, et~al.
  2017.
\newblock Visual genome: Connecting language and vision using crowdsourced
  dense image annotations.
\newblock \emph{International Journal of Computer Vision}, 123(1):32--73.

\bibitem[{Ma et~al.(2019)Ma, Lu, Wu, AlRegib, Kira, Socher, and
  Xiong}]{anonymous2019self-aware}
Chih-Yao Ma, Jiasen Lu, Zuxuan Wu, Ghassan AlRegib, Zsolt Kira, Richard Socher,
  and Caiming Xiong. 2019.
\newblock Self-monitoring navigation agent via auxiliary progress estimation.
\newblock In \emph{Proceedings of the International Conference on Learning
  Representations (ICLR)}.

\bibitem[{MacMahon et~al.(2006)MacMahon, Stankiewicz, and
  Kuipers}]{macmahon2006walk}
Matt MacMahon, Brian Stankiewicz, and Benjamin Kuipers. 2006.
\newblock Walk the talk: Connecting language, knowledge, and action in route
  instructions.
\newblock In \emph{Proceedings of the Conference on Artificial Intelligence
  (AAAI)}.

\bibitem[{Mei et~al.(2016)Mei, Bansal, and Walter}]{Mei16Instructions}
Hongyuan Mei, Mohit Bansal, and Matthew Walter. 2016.
\newblock Listen, attend, and walk: Neural mapping of navigational instructions
  to action sequences.
\newblock In \emph{Proceedings of the Conference on Artificial Intelligence
  (AAAI)}.

\bibitem[{Mirowski et~al.(2018)Mirowski, Grimes, Malinowski, Hermann, Anderson,
  Teplyashin, Simonyan, Zisserman, Hadsell et~al.}]{mirowski2018learning}
Piotr Mirowski, Matt Grimes, Mateusz Malinowski, Karl~Moritz Hermann, Keith
  Anderson, Denis Teplyashin, Karen Simonyan, Andrew Zisserman, Raia Hadsell,
  et~al. 2018.
\newblock Learning to navigate in cities without a map.
\newblock In \emph{Advances in Neural Information Processing Systems (NIPS)}.

\bibitem[{Mirowski et~al.(2017)Mirowski, Pascanu, Viola, Soyer, Ballard,
  Banino, Denil, Goroshin, Sifre, Kavukcuoglu et~al.}]{mirowski2016learning}
Piotr Mirowski, Razvan Pascanu, Fabio Viola, Hubert Soyer, Andrew~J Ballard,
  Andrea Banino, Misha Denil, Ross Goroshin, Laurent Sifre, Koray Kavukcuoglu,
  et~al. 2017.
\newblock Learning to navigate in complex environments.
\newblock In \emph{Proceedings of the International Conference on Learning
  Representations (ICLR)}.

\bibitem[{Misra et~al.(2018)Misra, Bennett, Blukis, Niklasson, Shatkhin, and
  Artzi}]{misra2018mapping}
Dipendra Misra, Andrew Bennett, Valts Blukis, Eyvind Niklasson, Max Shatkhin,
  and Yoav Artzi. 2018.
\newblock Mapping instructions to actions in 3d environments with visual goal
  prediction.
\newblock In \emph{Proceedings of the 2018 Conference on Empirical Methods in
  Natural Language Processing}, pages 2667--2678.

\bibitem[{Pennington et~al.(2014)Pennington, Socher, and
  Manning}]{pennington2014glove}
Jeffrey Pennington, Richard Socher, and Christopher Manning. 2014.
\newblock Glove: Global vectors for word representation.
\newblock In \emph{Proceedings of the 2014 conference on empirical methods in
  natural language processing (EMNLP)}, pages 1532--1543.

\bibitem[{Ren et~al.(2015)Ren, He, Girshick, and Sun}]{ren2015faster}
Shaoqing Ren, Kaiming He, Ross Girshick, and Jian Sun. 2015.
\newblock Faster r-cnn: Towards real-time object detection with region proposal
  networks.
\newblock In \emph{Advances in neural information processing systems}, pages
  91--99.

\bibitem[{Rohrbach et~al.(2018)Rohrbach, Hendricks, Burns, Darrell, and
  Saenko}]{rohrbach2018emnlp}
Anna Rohrbach, Lisa~Anne Hendricks, Kaylee Burns, Trevor Darrell, and Kate
  Saenko. 2018.
\newblock Object hallucination in image captioning.
\newblock In \emph{Proceedings of the Conference on Empirical Methods in
  Natural Language Processing (EMNLP)}.

\bibitem[{Tellex et~al.(2011)Tellex, Kollar, Dickerson, Walter, Banerjee,
  Teller, and Roy}]{tellex2011understanding}
Stefanie Tellex, Thomas Kollar, Steven Dickerson, Matthew~R Walter, Ashis~Gopal
  Banerjee, Seth~J Teller, and Nicholas Roy. 2011.
\newblock Understanding natural language commands for robotic navigation and
  mobile manipulation.
\newblock In \emph{AAAI}, volume~1, page~2.

\bibitem[{Thomason et~al.(2019)Thomason, Gordon, and
  Bisk}]{thomason2018shifting}
Jesse Thomason, Daniel Gordon, and Yonatan Bisk. 2019.
\newblock Shifting the baseline: Single modality performance on visual
  navigation \& qa.
\newblock In \emph{Conference of the North American Chapter of the Association
  for Computational Linguistics (NAACL)}.

\bibitem[{Vogel and Jurafsky(2010)}]{vogel2010learning}
Adam Vogel and Dan Jurafsky. 2010.
\newblock Learning to follow navigational directions.
\newblock In \emph{Proceedings of the 48th Annual Meeting of the Association
  for Computational Linguistics}, pages 806--814. Association for Computational
  Linguistics.

\bibitem[{Wang et~al.(2018)Wang, Xiong, Wang, and Yang~Wang}]{wang2018look}
Xin Wang, Wenhan Xiong, Hongmin Wang, and William Yang~Wang. 2018.
\newblock Look before you leap: Bridging model-free and model-based
  reinforcement learning for planned-ahead vision-and-language navigation.
\newblock In \emph{Proceedings of the European Conference on Computer Vision
  (ECCV)}, pages 37--53.

\bibitem[{Yang et~al.(2019)Yang, Wang, Farhadi, Gupta, and
  Mottaghi}]{yang2018visual}
Wei Yang, Xiaolong Wang, Ali Farhadi, Abhinav Gupta, and Roozbeh Mottaghi.
  2019.
\newblock Visual semantic navigation using scene priors.
\newblock In \emph{Proceedings of the International Conference on Learning
  Representations (ICLR)}.

\bibitem[{Zhu et~al.(2017)Zhu, Mottaghi, Kolve, Lim, Gupta, Fei-Fei, and
  Farhadi}]{zhu2017target}
Yuke Zhu, Roozbeh Mottaghi, Eric Kolve, Joseph~J Lim, Abhinav Gupta,
  Li~Fei-Fei, and Ali Farhadi. 2017.
\newblock Target-driven visual navigation in indoor scenes using deep
  reinforcement learning.
\newblock In \emph{Robotics and Automation (ICRA), 2017 IEEE International
  Conference on}, pages 3357--3364. IEEE.

\end{thebibliography}

\clearpage
\appendix

\section*{Supplementary material to ``Are You Looking? Grounding to Multiple Modalities in Vision-and-Language Navigation''}

\section{Details on the compared VLN models}

The Speaker-Follower (SF) model \cite{fried2018speaker} and the Self-Monitoring (SM) model \cite{anonymous2019self-aware} which we analyze both 
use sequence-to-sequence model \cite{cho2014properties} with attention \cite{bahdanau2014neural} as their base instruction-following agent. 
Both use an encoder LSTM \cite{hochreiter1997long} to represent the instruction text, and a decoder LSTM to predict actions sequentially. At each timestep, the decoder LSTM conditions on the action previously taken, a representation of the visual context at the agent's current location, and an attended representation of the encoded instruction. 

While at a high level these models are similar (at least in terms of the base sequence-to-sequence models -- both papers additionally develop techniques to select routes from these base models during search-based inference techniques, either using a separate language generation model in SF, or a progress-monitor in SM), they differ in the mechanism by which they combine representations of the text instruction and visual input. 
The SM uses a co-grounded attention mechanism, where both the visual attention on image features and the textual attention on the instruction words are generated based on previous decoder LSTM hidden state $h_{t-1}$, and then the attended visual and textual features are used as LSTM inputs to produce $h_t$. The SF model only uses attended visual features as LSTM inputs and then produces textual attention based on updated LSTM state $h_t$. Also, the visual attention weights are calculated with an MLP and batch-normalization in SM, while only a linear dot-product visual attention is used in SF. 
Empirically these differences produce large performance improvements for the SM model, which may contribute to the smaller gap between the SM model and its non-visual counterparts.

\section{Details on the training mechanisms}
\citet{anderson2018cvpr} compare two methods for training agents, which subsequent work on VLN has also used. These methods differ in whether they allow the agent to visit viewpoints which are not part of the true routes at training time. 
In the first training setup, \emph{teacher-forcing}, the agent visits each viewpoint in a given true route in sequence, and is supervised at each viewpoint with the action necessary to reach the next viewpoint in the true route. In the second training setup, \emph{student-forcing}, the agent takes actions by sampling from its predicted distribution at each timestep, which results in exploring viewpoints that are not part of the true routes. At each viewpoint, supervision is provided by an oracle that returns the action which would take the agent along the shortest path to the goal. 
Empirically, student-forcing works better in nearly all settings in Table~\ref{tab:vis_nonvis} (except for the non-visual version of the SF model), which is likely due to the fact that it reduces the discrepancy between training and testing, since it allows the agent to sample from its own prediction during training. Teacher-forcing works better for the non-visual version of the SF model, and we hypothesize that following the ground-truth routes during training allows the SF model to better preserve the geometric structures of the routes and match them to the instructions for the non-visual setting.

\section{Details on the object representation}
In our object representation, we use the top-150 detected objects (with the highest detection confidence) at each location in the environment. The detection results are obtained from a Faster R-CNN detector \cite{ren2015faster} pretrained on the Visual Genome dataset \cite{krishna2017visual}.

\end{document}